\documentclass[sigconf]{acmart}

\renewcommand\footnotetextcopyrightpermission[1]{} % removes footnote with conference information in first column
\makeatletter
\renewcommand\@formatdoi[1]{\ignorespaces}
\makeatother

\usepackage{booktabs} % For formal tables
\usepackage{amsmath}
\usepackage{placeins}
\usepackage{array,multirow,graphicx}
\usepackage{afterpage}

\usepackage{caption,subcaption}
\usepackage{tikz}
\usetikzlibrary{positioning,fit}
\usetikzlibrary{matrix}

\definecolor{darkgreen}{rgb}{0.0, 0.2, 0.13}
\definecolor{darkolivegreen}{rgb}{0.33, 0.42, 0.18}

\newcommand{\eg}{\textit{e.g.}\ }
\newcommand{\ie}{\textit{i.e.}\ }

\newcommand{\etal}{\textit{et al.}\ }

\AtBeginDocument{%
  \providecommand\BibTeX{{%
    \normalfont B\kern-0.5em{\scshape i\kern-0.25em b}\kern-0.8em\TeX}}}

\copyrightyear{X}
\acmYear{X}
\setcopyright{acmcopyright}
\acmBooktitle{X}
\acmPrice{X}
\acmDOI{X}
\acmISBN{X}

\begin{document}
%\title{Comparing DRL with NS/QD on the BipedalWalker benchmark}
%\title{Comparing MAP-Elites and Curiosity-driven A3C on the BipedalWalker benchmark\\
%Exploring the BipedalWalker benchmark with MAP-Elites and Curiosity-driven A3C
%}
\title{Ensemble Feature Extraction for Multi-Container Quality-Diversity Algorithms}

\author{Leo Cazenille}
\orcid{0000-0002-5893-9761}
\affiliation{%
  \institution{Ochanomizu University}
  %\streetaddress{P.O. Box 1212}
  \city{Tokyo} 
  \state{Japan} 
}
\email{leo.cazenille@gmail.com}

%\author{Anonymous Anonymous}
%\orcid{0000-0000-0000-0000}
%\affiliation{%
%  \institution{Anonymous Institute}
%  %\streetaddress{P.O. Box 1212}
%%  \city{Unknown} 
%%  \state{Earth} 
%  %\postcode{43017-6221}
%}
%\email{an.ony@mo.us}

%\titlenote{Produces the permission block, and copyright information}
%\subtitle{Extended Abstract}
%\subtitlenote{The full version of the author's guide is available as
%  \texttt{acmart.pdf} document}

\begin{abstract}
Quality-Diversity algorithms search for large collections of diverse and high-performing solutions, rather than just for a single solution like typical optimisation methods. They are specially adapted for multi-modal problems that can be solved in many different ways, such as complex reinforcement learning or robotics tasks. However, these approaches are highly dependent on the choice of feature descriptors (FDs) quantifying the similarity in behaviour of the solutions. While FDs usually needs to be hand-designed, recent studies have proposed ways to define them automatically by using feature extraction techniques, such as PCA or Auto-Encoders, to learn a representation of the problem from previously explored solutions. Here, we extend these approaches to more complex problems which cannot be efficiently explored by relying only on a single representation but require instead a set of diverse and complementary representations. We describe MC-AURORA, a Quality-Diversity approach that optimises simultaneously several collections of solutions, each with a different set of FDs, which are, in turn, defined automatically by an ensemble of modular auto-encoders. We show that this approach produces solutions that are more diverse than those produced by single-representation approaches. %We improve those results by tuning the number of containers considered, and training the modular auto-encoders according to a loss function with several diversity components.
\end{abstract}


\begin{CCSXML}
<ccs2012>
   <concept>
       <concept_id>10010147.10010178.10010205</concept_id>
       <concept_desc>Computing methodologies~Search methodologies</concept_desc>
       <concept_significance>500</concept_significance>
       </concept>
 </ccs2012>
\end{CCSXML}

\ccsdesc[500]{Computing methodologies~Search methodologies}

\keywords{Quality-Diversity algorithms, AURORA, Evolutionary Robotics, Ensemble representation learning, Auto-encoders, Deep Learning}

\fancyfoot{}

\maketitle
\thispagestyle{empty}

\begin{figure}[h]
\begin{center}
\includegraphics[width=0.49\textwidth]{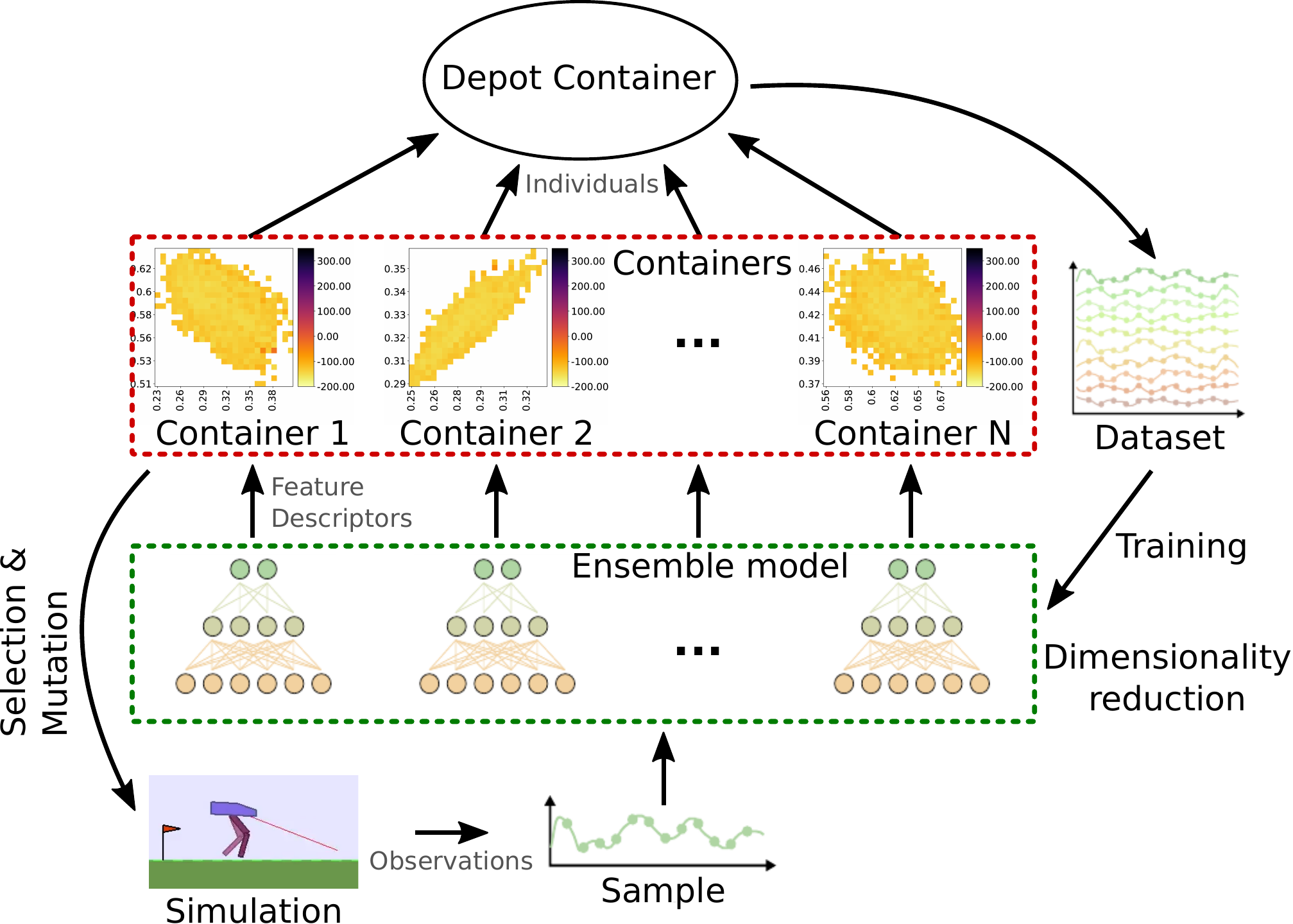}
\caption{Workflow of MC-AURORA: a Quality-Diversity algorithm explores the behavioural space of a given problem and stores the elite solutions into several containers, each with a different set of feature descriptors (FDs). These FDs are automatically discovered through an ensemble dimensionality reduction technique (\eg auto-encoders), that is periodically trained on the previously encountered solutions. The multi-container aspect of this method extends the AURORA~\cite{cully2019autonomous} algorithm to discover automatically an arbitrarily large number of diverse representations of the problem.}
\label{fig:workflow}
\end{center}
\end{figure}

\section{Introduction}
Designing deep neural network controllers for robotic tasks is still viewed as a complex problem, involving ill-defined objectives and dynamics, unknown and possibly hostile environments, and large experimental variability. One particular challenge is handling complex and uncertain situations that the robot would only be able to overcome by exhibiting behaviours it was not prepared and programmed for. For complex problems, these solutions may be heavily context-dependent, suggesting the use of approaches where the robot would select and adapt its behaviour depending on its environment. 
These approaches differ from typical search-based and learning-based methods, which usually aim to find one high-performing solution rather than collecting a large number of diverse controllers catering to various environmental contexts. 
%This differ from typical methods to train or search deep neural networks controllers, where the goal is usually to find one high-performing solution rather than collecting a large number of diverse controllers exhibiting various behaviours that can each cope with a different environmental context. 

This paradigm is realised by a recent family of optimisation algorithms named Quality-Diversity (QD) algorithms or Illumination algorithms~\cite{pugh2016quality,cully2018quality,chatzilygeroudis2020quality,cully2018quality}, such as Map-Elites~\cite{mouret2015illuminating} and Novelty Search with Local Competition~\cite{lehman2011evolving}.
Algorithms like Map-Elites~\cite{mouret2015illuminating} are grid-based, and regroup the explored solutions in a grid (or container~\cite{cully2018quality}) of elites. This produces sets of high-performing solutions that vary according to Feature Descriptors (FDs) represented as axes of the grid. These Feature Descriptors (also called Behavioural Descriptors if they are related to robot behaviours) define a (behavioural) characterisation of the problem and quantify how diverse each controller is from one another.

These algorithms were previously used to solve problems either by efficiently exploring deceptive search spaces~\cite{lehman2013effective}, or by identifying and exploiting the generated repertoire of solutions~\cite{mouret2015illuminating}.
In particular, they are particularly successful in the Evolutionary robotics ~\cite{cully2015robots,mouret2020evolving,duarte2018evolution} community, but they were also previously used in Reinforcement Learning~\cite{brych2020competitiveness,ecoffet2019go,gupta2020exploring}, for video games~\cite{fontaine2019covariance,alvarez2019empowering}, in Biology and Chemistry~\cite{verhellen2020illuminating,cazenille2019automatic,cazenille2019exploring}, and in problems typically solved with traditional optimisation algorithms~\cite{bruneton2019exploration}.

However, in most cases, the human expert must provide the set of FDs used by the QD algorithm to discriminate the different types of solutions. This choice of FDs may require prior knowledge of the task or a high level of expertise. It will also heavily impacts the efficiency of the search process, and results in collections of solutions that are catering to one specific range of interest. 

This problem was addressed through several recent studies that defined ways to automatically identify a set of FDs of a problem, without relying on \textit{a-priori} information. Notably, Cully proposed the AURORA ("AUtonomous RObots that Realize their Abilities") algorithm~\cite{cully2019autonomous} that combined QD with dimensionality reduction (DR) algorithms to perform feature extraction and discover automatically representations of the problem, which were then used as FDs. Indeed, DR methods, such as PCA~\cite{pearson1901liii} and Deep auto-encoders~\cite{le1986learning,bourlard1988auto} are classical tools in Machine Learning and representation learning to extract meaningful low-dimensional representations from high-dimensional data. Usually, a large number of samples are required to train these methods. However, this is not a problem for the AURORA algorithm, as QD algorithms typically explore a large number of solutions which can be, in turn, used to train the DR.

Prior to AURORA, the DeLeNoX was introduced by Liapis \etal~\cite{liapis2013transforming}. It involves two alternating phases: 1) one dimensionality reduction phase computing a diversity measure through a denoising auto-encoder; 2) a novelty-search phase based on this diversity measure. It can thus be seen as a diversity-only version of the AURORA algorithm, without consideration to solution quality.

AURORA also inspired other methods, like the TAXONS~\cite{paolo2020unsupervised} algorithm, that searches all potential outcomes of a given problem; or the DDE-Elites~\cite{gaier2020discovering} algorithm, that conjointly searches and refines a representation of the problem with the help of a Variational Auto-encoder model, and illuminates this search space with adaptive mutation operators.

However, these methods all rely on a single DR instance of the problem, biasing the search process towards a small subset of interesting solution niches in the problem space. AURORA and subsequent algorithms focused on relatively simple benchmark problems that could be characterised with a very small number of FDs (\eg two dimensions). Yet complex problems can be represented in a large (potentially infinite) number of representations, each capturing different aspects of the problem. This property has spurred the representation learning community to come up with DR techniques that can capture a large diversity of features in a problem, a topic known as "ensemble representation learning", involving a collection of various DR modules.

Here, we extend the AURORA algorithm to be able to scale to more complex problems with an arbitrarily large number of distinct representations to code as FDs. This is achieved by enabling AURORA to work with a collection of complementary QD containers (/grids) rather than a single container. Each of these containers has FDs that are defined by a distinct DR module, and DR modules are trained together in an ensemble model. We name this algorithm "\textbf{MC-AURORA}": Multi-Container AURORA (Fig.~\ref{fig:workflow}).

We test \textbf{MC-AURORA} to explore a typical Deep reinforcement problem: the OpenAI Gym BipedalWalker-v3 environment where a bipedal agent must locomote to the end of a slightly uneven randomly generated terrain. We showcase our approach with an ensemble of modular auto-encoders as a DR technique.

We demonstrate that \textbf{MC-AURORA} with several containers can find more diverse collections of solution than the original AURORA algorithm with only one container, and in a more reliable manner. We investigate how the number of containers used can influence the performance of the algorithm. Finally, We present several loss functions that can be applied to the training of the auto-encoders so that the representations found by DR are as different as possible to one another.

\section{Methods}

\subsection{AURORA}
The AURORA algorithm first evaluates a given amount of randomly initialised solutions to compile a dataset of their respective observations matrices. It is used to train the dimensionality reduction algorithm (DR), such as an Auto-Encoder, to learn of (low-dimensional, \eg 2D) latent representation of the observations (a Feature extraction task). This model will serve as a representation of the FD space for the QD container. As such, it can be used to compute the FD of an arbitrary evaluated solution by projecting the latter's observations into the latent space. AURORA proposes two DR methods of choice: Principal Component Analysis (PCA)~\cite{pearson1901liii} and Deep (Convolutional) Auto-encoders~\cite{le1986learning,bourlard1988auto}.

Then, AURORA behaves like normal QD algorithm. For each evaluation of the algorithm, 1) a solution is selected in the container, 2) it is varied/mutated, 3) evaluated to compute its fitness and FD (from the latent space projection), 4) and finally an attempt is made to place it into the container. The solutions stored into the container iteratively increase, which also increase the number of observations available to compile the training dataset. This method can work with any kind of containers~\cite{cully2018quality}, such as Grid, or Novelty Archives. For instance, the original AURORA paper used Novelty Archive containers with exclusive $\epsilon$-dominance and a self-adaptive novelty threshold parameter.

There are then two strategies for training. With the \textbf{Pre-training} strategy, the DR model is never re-trained again after the initial training. 
With the \textbf{Online} strategy, the DR algorithm is periodically re-trained over this dataset (\eg every $k$ iterations, or with a period that decreases exponentially, or even every time $k$ new observations vectors are added). After it is done, the FD of all solutions of the container must be recomputed to use the newly re-trained DR model: this can be done by emptying first the container of all solutions, then recomputing the FD, and finally adding back the solutions one at a time. This may translate into large changes inside the container, as some solutions with previously distinct FD may in turn have very similar FDs after the FD recomputation. In this case, only the best ones are kept, in accordance with the competition mechanism of the QD algorithm.

\subsection{Multi-Container AURORA}
The original AURORA relies on only one DR model to project observations into their latent representations. As such, it may heavily bias its search process to favour the niches in the FD space that are captured through its DR model, while ignoring others. While this may not be a problem for simpler FD space, it poses the problem of scalability for more complex problems. Moreover, only relying on one model will increase the variability of its accuracy to represent a particular problem depending on its initial conditions (\eg initial weights for a neural network).

Here, we describe a variant of the AURORA algorithm for Multi-container settings. A Multi-container method for Quality-Diversity algorithms ("MCQD") was originally presented in Doncieux \etal~\cite{doncieux2018open}, where a collection of containers of different types could be used concurrently during a single search process. This design is reminiscent of the Island models in Evolutionary Computation, where several populations of solutions are optimised concurrently~\cite{skolicki2005analysis}.

In this paper, we show that MCQD and AURORA can be combined to use concurrently a collection of containers, each with a distinct (low-dimensional, \eg 2D) FD space that represents complementary behavioural characterisations and matches different niches in the problem space. This is made possible by using a collection of DR models to represent the FD space of each container.

The workflow of the \textbf{MC-AURORA} algorithm is presented in Fig.~\ref{fig:workflow}. It is similar to the AURORA algorithm, with a few key differences. First, all containers are initialised with the same collection of randomly initialised solutions. This may translate into redundancies of solutions across containers where a solution can be stored in several containers at the same time, especially if the FD spaces of each container are similar to one another. Second, during the search process, if one solution is added to one of the controllers, it is also stored inside a "depot container", that contain solutions found for all containers. This "depot container" will then serve to compile the training dataset used to train the collection of DR models.

To modulate the number of redundancies across all containers, we consider two strategies.
In the \textbf{Shared solutions} strategy, each time a new solution is selected, mutated, and evaluated, an attempt is made to add it to every container. With this strategy, good solutions will tend to propagate to all container, reducing the overall diversity of solutions but providing a potential competitive advantage in term of performance. In the \textbf{Non-shared solutions} strategy, the search process only focuses on one container at a time: each time a new solution is selected, mutated, and evaluated, an attempt is made to add it to the focused container. Then the focus changes to a different container, with each container having the same budget of evaluations.

For reasons of simplicity, we will use Grid containers instead of the Novelty Archives used in the original AURORA paper. However, our approach could be used with any kind of container.

%~\cite{farrell2020autoencoder}

%\subsection{Dimensionality reduction for feature extraction}
%\TODO{}

\subsection{Ensemble training of Modular Auto-Encoders}

In the field of representation learning, methods exist to train an \textbf{ensemble} model composed of several independent modules that are trained on the same dataset~\cite{bolon2019ensembles}. Ensemble models usually combine the outputs of a collection of (potentially diverse) models to obtain enriched output space while controlling variance~\cite{hastie2009elements}.
Similarly, Reeve \etal~\cite{reeve2015modular,reeve2018diversity} presented a modular method to train concurrently a diverse ensemble of independent auto-encoders. It allows extracting several distinct sub-sets of features resulting in an enriched latent space representation, with a trade-off between accuracy (reconstruction) and diversity. This method involves the use of a loss function with a diversity component (to ensure that each module is different), which will be presented in the following section.

\newcommand{\AEE}{\ensuremath{\mathbf{E}}}
\newcommand{\AED}{\ensuremath{\mathbf{D}}}
\newcommand{\AEx}{\ensuremath{\mathbf{x}}}
\newcommand{\norm}[1]{\left\lVert#1\right\rVert}
We consider the following notation: an auto-encoder is composed of two modules: an encoder $E: \mathcal{X} \rightarrow \mathcal{Z}$ and a decoder $D: \mathcal{Z} \rightarrow \mathcal{X}$. Let $\mathcal{X} = \mathbb{R}^{n\times t}$ be the observation space, with $n$ the number of observations per time steps of simulation and $t$ the number of time steps. We set $\mathcal{Z} = \mathbb{R}^d$ as the latent space of the auto-encoder. Let $y = D \circ E(x)$ the output of the auto-encoder for $x \in \mathcal{X}$.

For a training batch $\AEx$ of size $B$, we denote $x^{(i)}$ the $i$th data sample and $z^{(i)} = E(x^{(i)})$ its code.
We use the mean squared error (MSE) as the reconstruction loss function:
\[ \mathcal{L}_{recons}(X) = \frac{1}{B} \sum_{i=1}^B \norm{D \circ E(x^{(i)}) - x^{(i)}}_{2}^{2} \]

As defined in~\cite{reeve2015modular}, a modular auto-encoder is an ensemble model $\mathcal{W}=\{(E_i, D_i)\}_{i=1}^{M}$ regrouping $M$ auto-encoder modules $(E_i, D_i)$.
The same paper defines a loss function designed to increase the diversity of the auto-encoder modules, and based on the reconstructed outputs of each module. It is defined as:
\begin{equation}
\mathcal{L}_{outputs}(\mathcal{W}, X) = \frac{1}{B} \frac{1}{M} \sum_{i=1}^{B} \sum_{j=1}^{M} \norm{D_j \circ E_j(x^{(i)} - \frac{1}{M} \sum_{k=1}^{M} D_j \circ E_j(x^{(k)}  ))}_2^2
\end{equation}

This loss function is combined with the reconstruction loss as follows:
\begin{equation}
\mathcal{L}(\mathcal{W}, X) = \mathcal{L}_{recons}(\mathcal{W}, X) - \lambda \mathcal{L}_{outputs}(\mathcal{W}, X)
\label{eq:loss}
\end{equation}
where $\lambda$ is a weighting parameter to define the importance of diversity compared to reconstruction.

In this paper, we also consider alternative loss functions to enforce diversity in modular Auto-Encoder. We postulate that diversity in latent representations can be obtained through loss functions based on the latent representation rather than reconstructed outputs. We want to constrain the latent representation of each container to behave as differently as possible from one another. This can be achieved through covariance-based statistics of either all the regrouped latent spaces of all containers or by comparing each latent space with one another. As such we define two loss functions as follows:

We define the following loss function computed using the mean magnitude of the covariance matrix (inspired from~\cite{ladjal2019pca}):
\begin{equation}
\mathcal{L}_{cov}(\mathcal{W}, X) = \sum_{i=1}^{d} \sum_{j=1, i\neq j}^d | cov(X)_{i,j} |
\end{equation}
where $cov(X)_{i,j}$ is the $i,j$ element of the covariance matrix of $X$. This loss function can be used in Eq.~\ref{eq:loss} instead of $\mathcal{L}_{outputs}(\mathcal{W}, X)$

We take inspiration the Correlation Matrix Distance defined in~\cite{herdin2005correlation}, and define the following loss function:
We set $R_1, ..., R_m$ the correlation matrices between each elements of the respective auto-encoders modules $\{(E_i, D_i)\}$.
\begin{equation}
d_{corr}(H_1, H_2) = 1 - \frac{tr\{H_1 H_2\}}{ \norm{H_1}_f \norm{H_2}_f } \in [0, 1]
\end{equation}
where $\norm{\cdot}_f$ is the Frobenius norm, and $tr\{\cdot\}$ the matrix trace.
\begin{equation}
\mathcal{L}_{cmd}(\mathcal{W}, X) = \sum_{i=1}^{d} \sum_{j=1, i\neq j}^d d_{corr}(R_i, R_j)
\end{equation}

%Histogram loss:~\cite{ustinova2016learning}

\subsection{Post-processing}
%Quantile tranform to uniform distrib from Scikit-learn (QuantileTransformer class):~\cite{scikit-learn}
Artificial Neural Network models trained through a gradient-based approach tend to favour data following normal distributions. This is also the case for the latent representations of auto-encoders. However, most Quality-Diversity algorithms use containers (including Grids and Novelty Archives) that are designed for FD following a uniform distribution (\ie uniform bins for Grids, constant novelty thresholds for Archives). As such, containers using the latent representation to define their FD space may have an artificially lowered coverage: Grids will have a lot of emptied bins and Archives will be artificially more sparse. Moreover, the FD space will be denser the closer to the distribution mean, resulting in non-uniform levels of competition between elites over the entire container.

Here, we propose to alleviate these problems by transforming the latent distributions to a uniform distribution ($\in [0, 1]$) by using Quantile Transformations~\cite{scikit-learn}, a methodology commonly used in Machine Learning to normalise data-sets.

The \textbf{MC-AURORA} algorithm was coded using the Python Quality-Diversity library QDpy~\cite{qdpy} version $0.1.2.1$.
All deep neural networks models are implemented with the PyTorch~\cite{pytorch} library version $1.7.1$.
All source codes are available at:~\url{https://github.com/leo-cazenille/multiAE-ME}.

\section{Experimental Validation}
All experimental cases considered are listed in Table~\ref{tab:cases}, and described in the following subsections. All tested algorithms are repeated 20 times and use the hyper-parameters listed in Table~\ref{tab:hyperparameters}.
All containers of all cases are initialised with a collection of 10000 solutions (with genomes following a random uniform distribution). This same collection is also used as the initial training data-set for the auto-encoders.

\begin{table*}[h]
\begin{center}
% TODO UPDATE !!!
\resizebox{1.00\textwidth}{!}{%
\begin{tabular}{p{0.3cm} l l l l l l l l l}
\hline
& Case Name & FD type & Shared solutions & Training type & Loss & Nr. of grids & Grids shape \\
\hline
\parbox[t]{2mm}{\multirow{6}{*}{\rotatebox[origin=c]{90}{Base}}}
 & hardcoded-4 & Hardcoded & Yes & None & None & 4 & 25x25  \\
 & hardcoded-4-ns & Hardcoded & No & None & None & 4 & 25x25 \\
 & pt-reco-4 & AE & Yes & Pre-trained & Reconstruction & 4 & 25x25 \\
 & reco-4 & AE & Yes &  Online & Reconstruction & 4 & 25x25 \\
 & qt-reco-4 & AE+QT & Yes & Online & Reconstruction & 4 & 25x25 \\
 & qt-reco-4-ns & AE+QT & No & Online & Reconstruction & 4 & 25x25 \\
\hline
\parbox[t]{2mm}{\multirow{5}{*}{\rotatebox[origin=c]{90}{Nr. of Grids}}}
 & hardcoded-1 & Hardcoded & - & None & None & 1 & 50x50 \\
 & qt-reco-1 & AE+QT & - & Online & Reconstruction & 1 & 50x50 \\
 & qt-reco-6-ns & AE+QT & No & Online & Reconstruction & 6 & 5*20x20 + 1*20x25 \\
 & qt-reco-9-ns & AE+QT & No & Online & Reconstruction & 9 & 8*17x16 + 1*18x18 \\
 & qt-reco-25-ns & AE+QT & No & Online & Reconstruction & 25 & 10x10 \\
\hline
\parbox[t]{2mm}{\multirow{4}{*}{\rotatebox[origin=c]{90}{Loss}}}
 & qt-outputs-4-ns & AE+QT & No & Online & Reconstruction+1*Outputs & 4 & 25x25 \\
 & qt-covmin-4-ns & AE+QT & No & Online & Reconstruction+1*COV & 4 & 25x25 \\
 & qt-covmax-4-ns & AE+QT & No & Online & Reconstruction-1*COV & 4 & 25x25 \\
 & qt-cmd-4-ns & AE+QT & No & Online & Reconstruction+1*CMD & 4 & 25x25 \\
\hline
\end{tabular}
}
\caption{List of all experimental cases. All cases have the same budget of 2500 bins over all their containers.}
\label{tab:cases}
\end{center}
\end{table*}

\begin{table}[h]
\begin{center}
\resizebox{0.49\textwidth}{!}{%
\begin{tabular}{p{0.3cm} p{3cm} p{6cm}}
\hline
& \textbf{Parameter Name} & \textbf{Value} \\
\hline

\parbox[t]{2mm}{\multirow{4}{*}{\rotatebox[origin=c]{90}{Simulation}}}
 & Episodes per eval. & 5\\
 & Max episode length & 300 (instead of the default 3000)\\
 & Controller topology & MLP 2 hidden layers 40x40, tanh activation\\
 & Prob. dimensionality & 2804\\
\hline

\parbox[t]{2mm}{\multirow{8}{*}{\rotatebox[origin=c]{90}{Illumination}}}
 & Grid total nr. of bins & 2500\\
 & Evaluation budget & 100000 (excl. initialisation)\\
 & Initialisation budget & 10000 (random uniform)\\
 & Batch size & 1000 Evaluations\\
 & Selection & Curiosity roulette (Score-proportionate)~\cite{cully2018quality} \\
 & Mutation & Polynomial bounded~\cite{mouret2015illuminating,deb2002fast} \\
 & Mutation probability & 0.1 \\
 & ETA~\cite{deb2002fast} & 20 (Mutation Crowding degree) \\
\hline

\parbox[t]{2mm}{\multirow{16}{*}{\rotatebox[origin=c]{90}{Feature-extraction}}}
 & Encoder topology & Two Conv1d (kernel size=3) with batchnorm and ELU activation, two dense of 5 then 2 neurons with dropouts (0.2) and Sigmoid activation\\
 & Decoder topology & Two dense of 2 then 5 neurons with dropouts (0.2) and ELU activation, two ConvTranspose1d (kernel size=3) with MaxUnpool1d, batchnorm and ELU activation, one dense with Sigmoid activation \\
 & Weight initialization & Xavier uniform \\
 & Training period & 5000 solutions added to the depot container\\
 & Optimiser & Adam\\
 & Nr. of epochs & 200\\
 & Learning rate & 0.1\\
 & Batch size & 1024\\
 & Validation split & 0.25\\
 & Diversity coeff. & 1.0\\
\hline
\end{tabular}
}
\caption{Hyper-parameters used for all experiments.}
\label{tab:hyperparameters}
\end{center}
\end{table}

\begin{figure}[h]
\begin{center}
\includegraphics[width=0.47\textwidth]{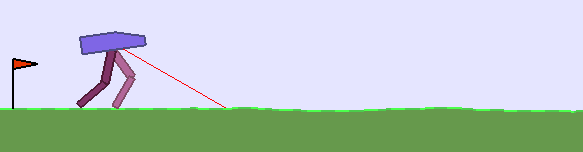}
\caption{Example frame of the OpenAI Gym~\cite{brockman2016openai} BipedalWalker-v3 environment where a robot with two legs and equipped with a lidar aims to move forward as far as possible in a stochastic environment.}
\label{fig:bipedal}
\end{center}
\end{figure}

\subsection{The BipedalWalker-v3 environment}

We test the performance of the \textbf{MC-AURORA} algorithm over the OpenAI Gym~\cite{brockman2016openai} BipedalWalker-v3 environment, a classical benchmark in the Reinforcement Learning and Quality-Diversity literature (Fig.~\ref{fig:bipedal}). This benchmark is more complex than those originally presented in the AURORA paper~\cite{cully2019autonomous}, with behaviours that can be characterised in numerous complementary ways, making it relevant as a benchmark for this study.

The BipedalWalker-v3~\footnote{\url{https://github.com/openai/gym/blob/master/gym/envs/box2d/bipedal_walker.py}} environment consists of a two-legged walking simulated agent that must locomote to the end of a slightly uneven randomly generated terrain in a limited amount of time. The agent is equipped with four torque-controlled motor driven joints. The 24 parameter state space includes position, hull angle, jumping, hip and knee joint angles/velocities, and a ten-point lidar range finder. The continuous action space is the four motor controlled torque values. The objective of the agent is to locomote the agent to the end of the environment. A positive reward is awarded for moving towards the goal, with an added bonus related to the stability of the hull. Motor efficiency is awarded a negative reward for the use of torque while a negative reward of -100 is awarded when the agent falls. 

Note that, we use a \textbf{simplified version} of the BipedalWalker-v3 benchmark with \textbf{only 300 steps per episode, instead of the default 3000}. Observations are also averaged over 30 time-steps (so there are 10 averaged state values per evaluation), and each evaluation is computed over 5 episodes. Each evaluation has thus $10 * 24 * 5 = 1200$ observations, which is lower than for the standard version ($180000$ observations per evaluation), making it simpler to model with an auto-encoder. The best fitness found are also around nine times lower ($40$) than the maximum obtainable of $350$.
The agent is controlled with an MLP controlled with a topology from the literature~\footnote{\url{http://blog.otoro.net/2017/11/12/evolving-stable-strategies/}}, as defined in Table~\ref{tab:hyperparameters}.

Benchmarking Quality-Diversity algorithms with a BipedalWalker environment was pioneered by the work of Justensen \etal~\cite{justesen2019map} where an adaptive sampling methodology was used to handle the stochasticity of this benchmark. Here, we use a simpler scheme to handle stochasticity, with fitness and FD averaged explicitly over the results of five episodes. The subsequent work of Gupta \etal~\cite{gupta2020exploring} used this approach to compare the performance of MAP-Elites and policy gradients algorithms trained with intrinsic curiosity loss.
This last work also presented the following set of human-designed FD (averaged over one episode) used to characterise the behaviour of the agent: 
% TODO FROM GECCO2020
\textit{Distance} is the agent's position relative to the goal, \textit{Hull angle} is the angle of the body of the agent, \textit{Torque} is the force applied to the agent's hip and knee joints, \textit{Jump} describes when both legs simultaneously are contact-less with the environment, \textit{Hip and knee angles} describe the agent's leg joint angles, \textit{Hip and knee speeds} describes the agent's leg joint angle speeds. 

As a reference experimental case, we test the performance of MAP-Elites over this benchmark with a similar set of hand-designed FDs as~\cite{gupta2020exploring}, with the exclusion of \textbf{Hip and knee speeds} (for the sake of simplicity). We consider three experimental cases with these FDs. First, \textbf{hardcoded-4} involves four bi-dimensional grids with shared solutions and with the following FDs pairs: (1) Distance vs. Hull Angle (directly correlated to the extrinsic reward); (2): Torque vs. Jump (impacts the overall efficiency of the agent); (3) and (4): joint angles between hips and knees for respectively the first and second legs. Second, \textbf{hardcoded-4-ns} is a similar case, but with non-shared solutions. Lastly, \textbf{hardcoded-1} only has one grid with the (1) pair of FD.

\afterpage{
\begin{figure*}[h!]
\begin{center}
\includegraphics[width=0.90\textwidth]{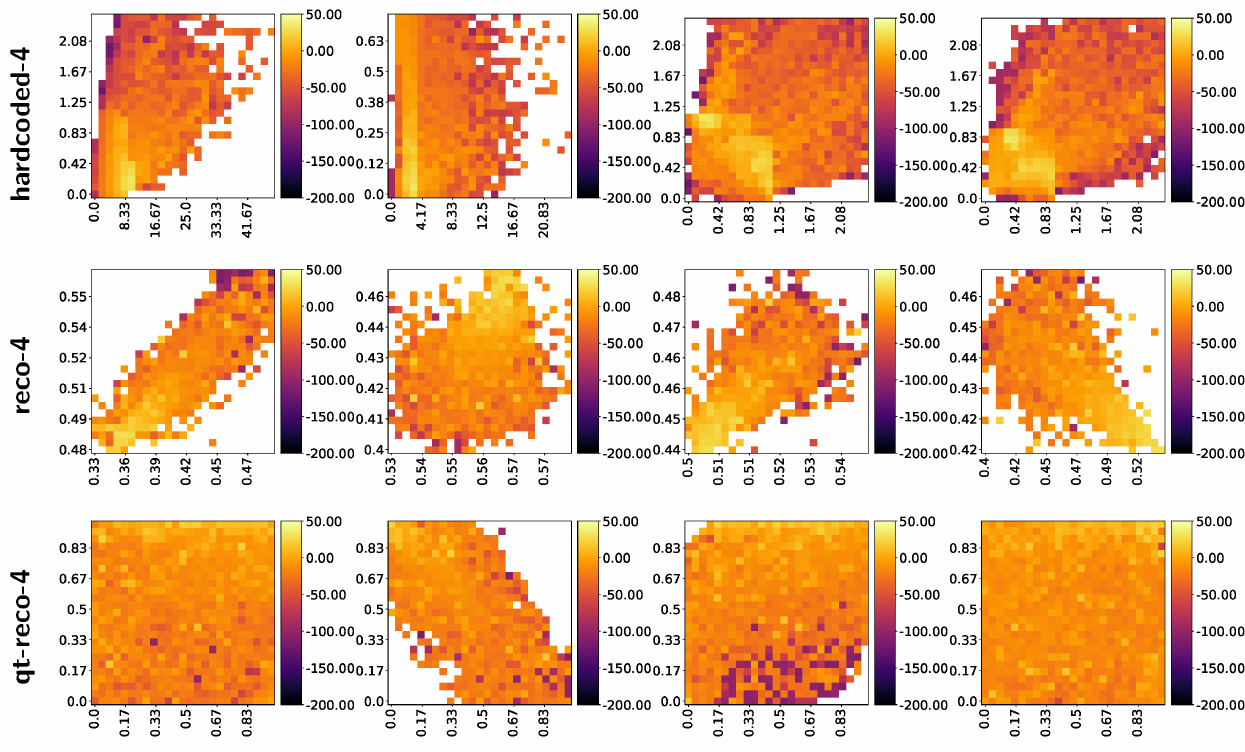}
\caption{Instances of the four grids of solutions found by three representative cases: \textbf{hardcoded-4} where the FD are hand-crafted by the expert, \textbf{reco-4} where the FD are the latent space of an Auto-Encoder, \textbf{qt-reco-4} where we post-process this latent space with a quantile transformation~\cite{scikit-learn} to follow a $[0, 1]$ uniform distribution. All cases have the same budget of 2500 grid bins.}
\label{fig:res1}
\end{center}
\end{figure*}
\begin{figure*}[h!]
\begin{center}
\includegraphics[width=0.32\textwidth]{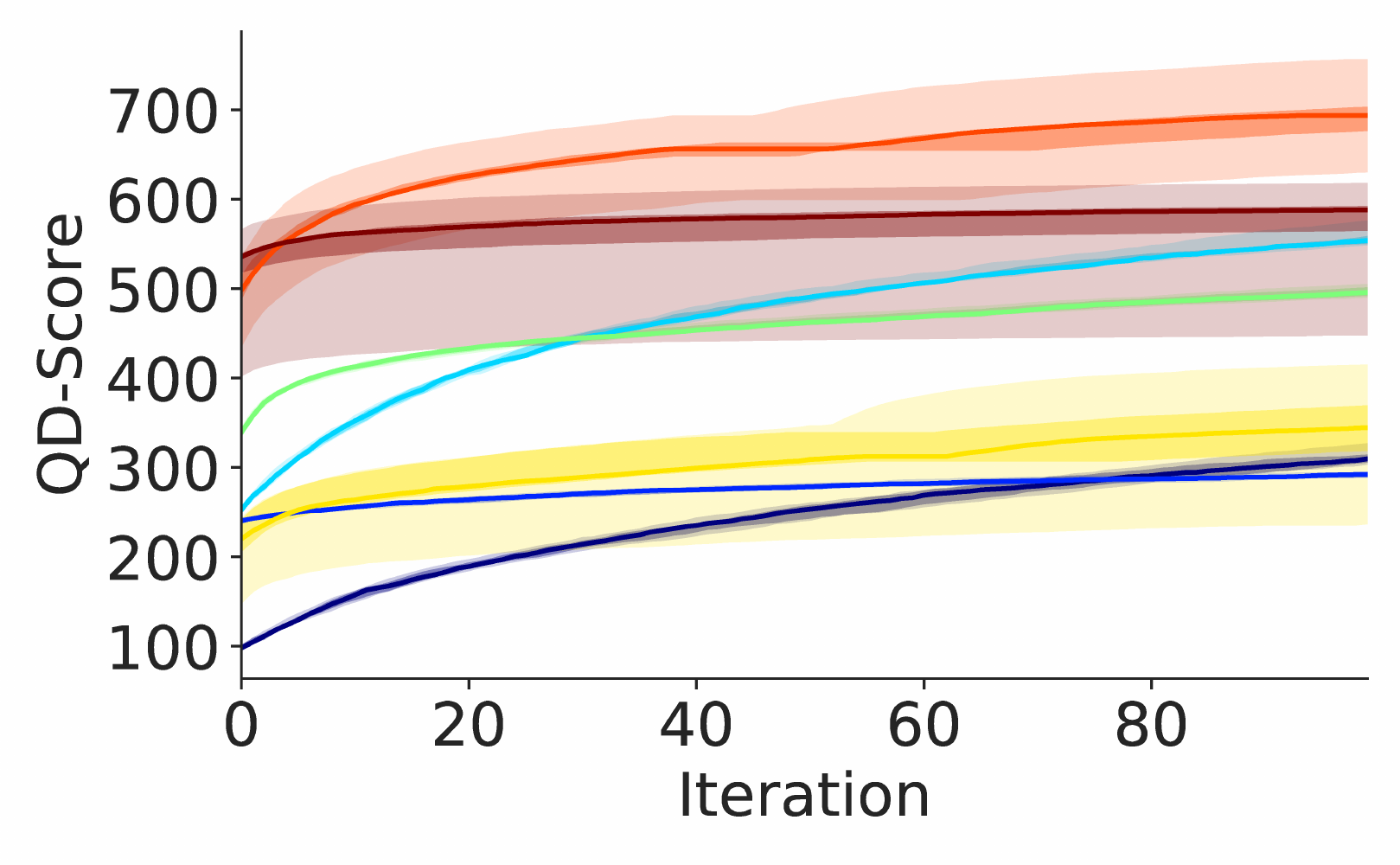}%
\includegraphics[width=0.32\textwidth]{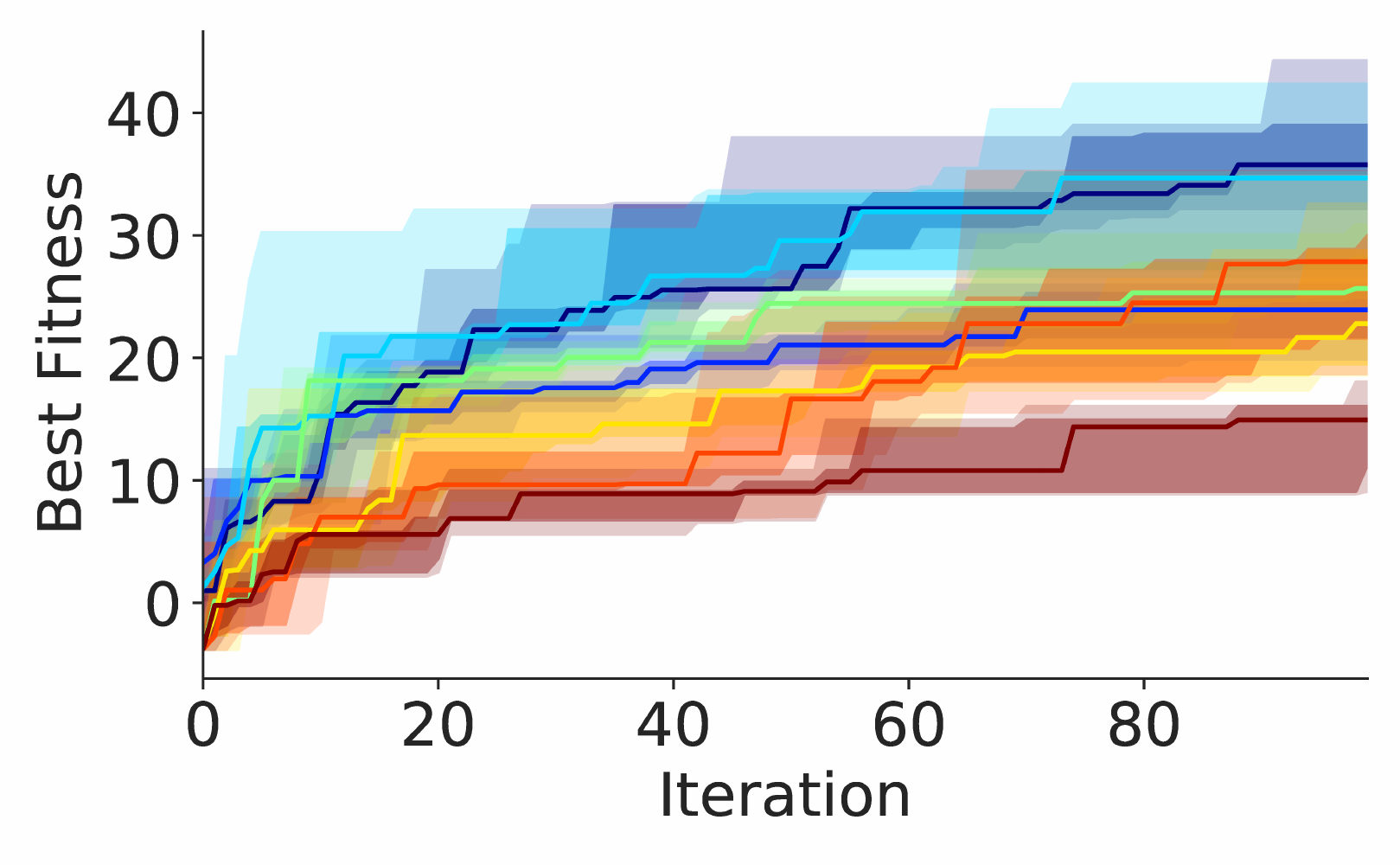}%
\includegraphics[width=0.32\textwidth]{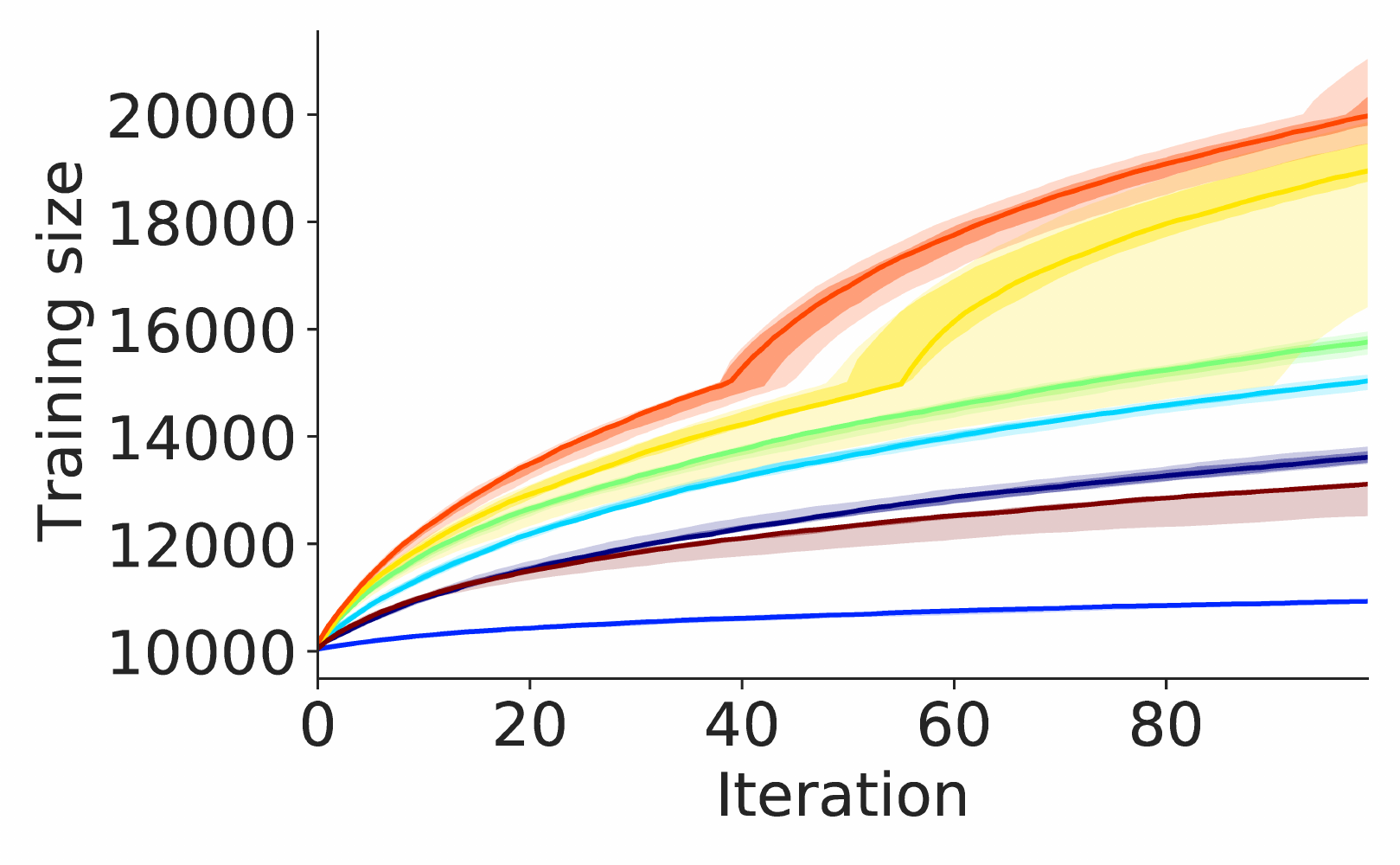}

\includegraphics[width=0.32\textwidth]{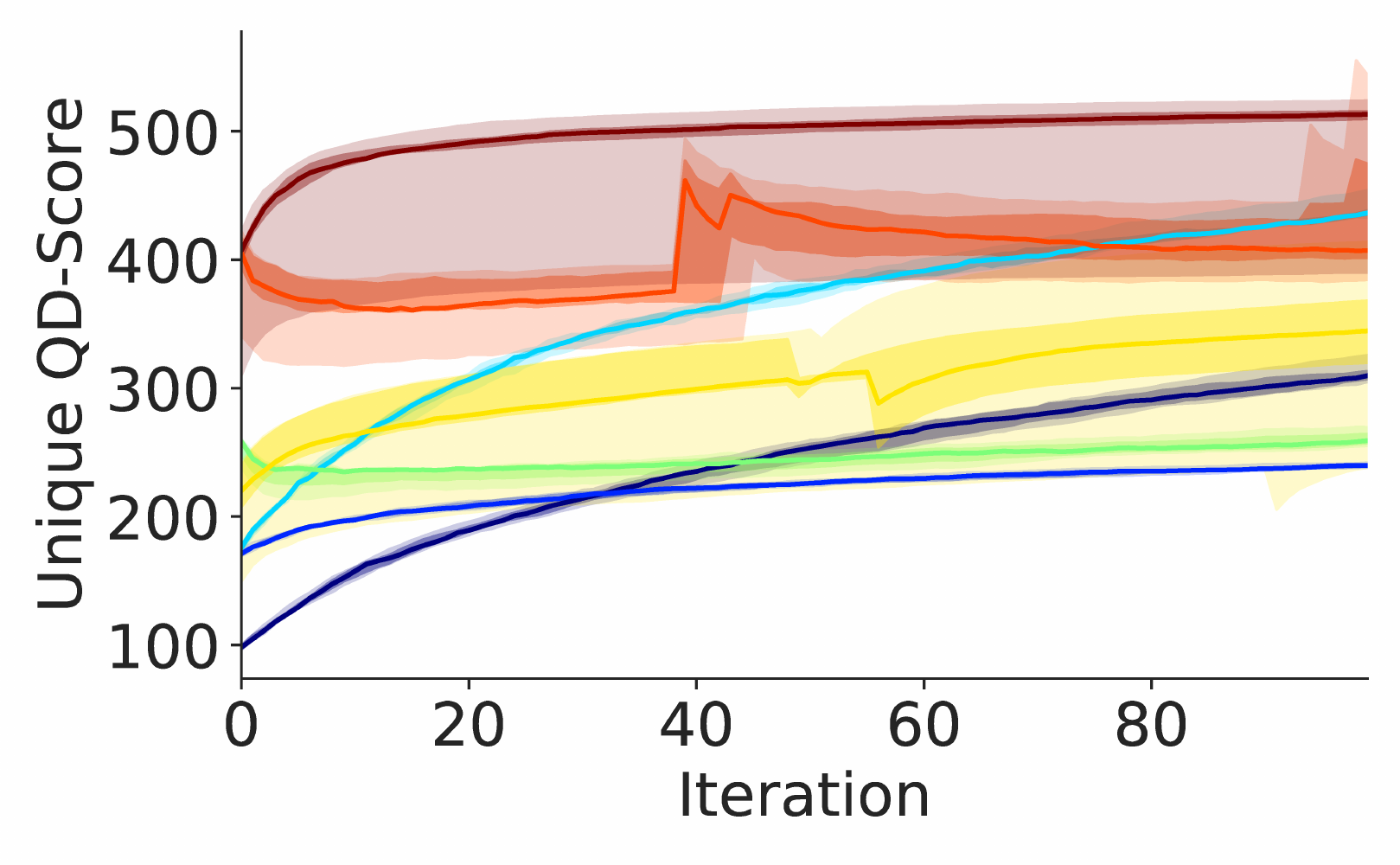}%
\includegraphics[width=0.32\textwidth]{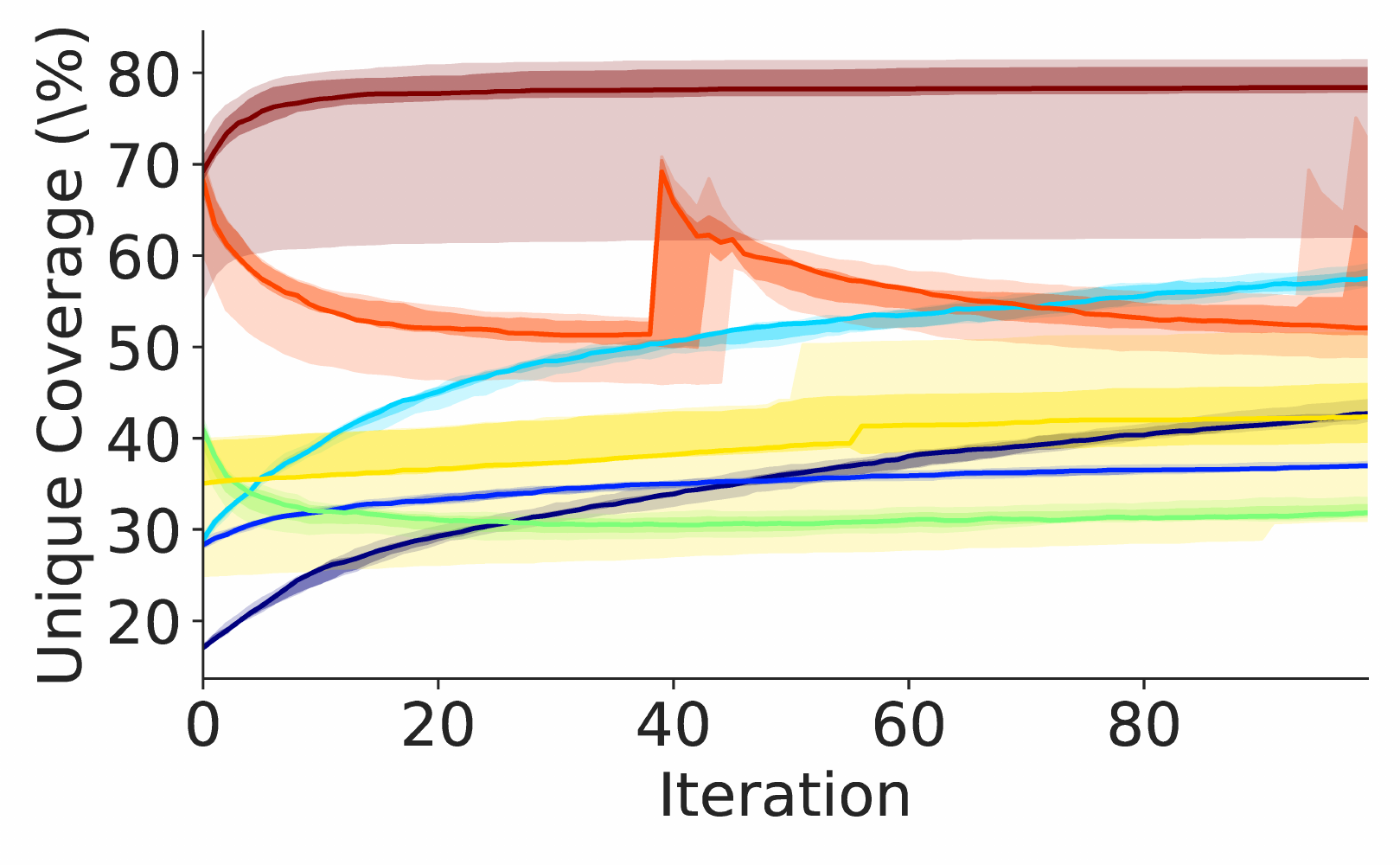}%
\includegraphics[width=0.20\textwidth]{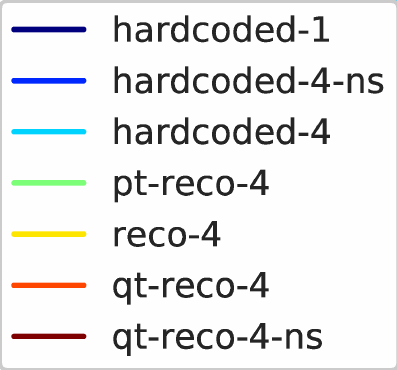}
\caption{Comparison of seven representative cases across the following metrics: (\textbf{top}) the QD-Score and best-ever fitness of solutions stored in all containers; (\textbf{top right}) depot container (auto-encoders training dataset) size; (\textbf{bottom}) the QD-Score and coverage of all unique solution stored in all containers (\ie without taking in account redundant solutions). Solid lines represent the mean of values across 20 runs. Light and dark shades correspond respectively to min-max values and 25th-75th quantiles.}
\label{fig:resPlots}
\end{center}
\end{figure*}
}

\subsection{Metrics}
Studied experimental cases are quantitatively compared through the following set of metrics. First, we assess the general performance and diversity of the solutions found at the last iteration of the algorithms, through the coverage (nr. of occupied bins), QD-score (sum of fitness normalised to $[0, 1]$ of all solutions in the containers), and best-fitness metrics.

Good solutions tend to propagate through several containers at the same time (including the initial 10000 solutions), especially in shared solutions cases. It may lead to a degree of solution redundancy across several containers. As such, two versions of the previously-defined metrics are considered: the \textbf{base} version is computed over all solutions found in all containers; the \textbf{unique} version takes into account the redundancy of solutions across several containers, and only count redundant solutions once.
We also assess the proportion of redundant solutions through a Container Redundancy metric, defined as:$1 - R/S$ with R the number of redundant solutions and $S=2500$ the sum of the capacity (\ie the number of bins) of each container.

We postulate that this redundancy, and also the general performance and diversity is highly dependant on the correlation between the FD used for all containers. As such, we define a metric termed "FD absolute correlation" computed as the mean magnitude (absolute value) of the correlation matrix components of the FD or all containers. A high score of this metric would translate into FD space that characterise highly similar behaviours across all containers.

\begin{table*}[h]
\begin{center}
\resizebox{0.95\textwidth}{!}{%
% TODO UPDATE !!!
\begin{tabular}{llllllll}
\hline
                 & Unique QD-Score      & Unique Coverage (\%)   & QD-Score             & Coverage (\%)      & Best Fitness       & FD Abs. Corr.     & Cont. Redundancy   \\
\hline
 hardcoded-4     & $439.360 \pm 7.471$  & $57.680 \pm 0.900$     & $557.496 \pm 8.885$  & $72.208 \pm 1.142$ & $33.229 \pm 5.781$ & $0.575 \pm 0.009$ & $0.368 \pm 0.006$  \\
 hardcoded-4-ns  & $239.614 \pm 2.857$  & $36.968 \pm 0.398$     & $291.560 \pm 2.817$  & $45.160 \pm 0.390$ & $25.353 \pm 4.828$ & $0.660 \pm 0.009$ & $0.352 \pm 0.001$  \\
 pt-reco-4       & $261.242 \pm 5.893$  & $32.240 \pm 0.833$     & $497.007 \pm 4.894$  & $59.320 \pm 0.770$ & $26.905 \pm 2.866$ & $0.277 \pm 0.045$ & $0.723 \pm 0.014$  \\
 reco-4          & $314.882 \pm 44.915$ & $39.232 \pm 5.106$     & $314.882 \pm 44.915$ & $39.232 \pm 5.106$ & $24.614 \pm 4.348$ & $0.483 \pm 0.323$ & $0.684 \pm 0.006$  \\
 qt-reco-4       & $473.456 \pm 67.374$ & $60.552 \pm 8.324$     & $722.727 \pm 68.967$ & $90.152 \pm 7.558$ & $23.319 \pm 5.246$ & $0.328 \pm 0.071$ & $0.553 \pm 0.090$  \\
 qt-reco-4-ns    & $505.298 \pm 46.870$ & $77.248 \pm 5.879$     & $574.096 \pm 57.843$ & $88.112 \pm 7.065$ & $17.078 \pm 3.104$ & $0.302 \pm 0.058$ & $0.241 \pm 0.027$  \\
\hline
 hardcoded-1     & $320.762 \pm 16.747$ & $43.832 \pm 2.551$     & $320.762 \pm 16.747$ & $43.832 \pm 2.551$ & $34.255 \pm 5.490$ & $0.581 \pm 0.029$ & $0.000 \pm 0.000$  \\
 qt-reco-1       & $602.999 \pm 93.807$ & $94.856 \pm 8.113$     & $602.999 \pm 93.807$ & $94.856 \pm 8.113$ & $14.067 \pm 4.103$ & $0.072 \pm 0.127$ & $0.000 \pm 0.000$  \\
 qt-reco-6-ns    & $590.995 \pm 35.041$ & $83.568 \pm 4.252$     & $631.452 \pm 44.386$ & $89.512 \pm 5.276$ & $19.206 \pm 6.155$ & $0.268 \pm 0.019$ & $0.130 \pm 0.013$  \\
 qt-reco-9-ns    & $604.357 \pm 30.788$ & $78.176 \pm 3.370$     & $606.429 \pm 37.204$ & $89.976 \pm 3.692$ & $23.998 \pm 4.371$ & $0.282 \pm 0.027$ & $0.217 \pm 0.045$  \\
 qt-reco-25-ns   & $446.305 \pm 9.096$  & $63.872 \pm 1.240$     & $653.876 \pm 10.931$ & $92.440 \pm 1.139$ & $20.672 \pm 4.173$ & $0.267 \pm 0.017$ & $0.468 \pm 0.010$  \\
\hline
 qt-outputs-4-ns & $538.816 \pm 24.450$ & $80.120 \pm 4.047$     & $598.385 \pm 32.554$ & $90.728 \pm 5.398$ & $19.733 \pm 3.155$ & $0.262 \pm 0.012$ & $0.225 \pm 0.019$  \\
 qt-covmin-4-ns  & $507.327 \pm 42.212$ & $77.280 \pm 5.658$     & $572.712 \pm 53.720$ & $87.368 \pm 7.125$ & $16.968 \pm 5.626$ & $0.287 \pm 0.054$ & $0.293 \pm 0.022$  \\
 qt-covmax-4-ns  & $528.666 \pm 30.037$ & $79.056 \pm 5.066$     & $571.542 \pm 32.726$ & $87.352 \pm 5.813$ & $17.850 \pm 3.750$ & $0.244 \pm 0.048$ & $0.192 \pm 0.024$  \\
 qt-cmd-4-ns     & $539.779 \pm 22.647$ & $79.784 \pm 2.878$     & $597.727 \pm 33.805$ & $90.096 \pm 4.073$ & $19.144 \pm 3.546$ & $0.252 \pm 0.034$ & $0.224 \pm 0.019$  \\
\hline
\end{tabular}

}
\caption{Mean statistics of all studied cases at the last iteration (100th) over 20 runs. The numbers following the symbol $\pm$ represent the standard deviation. The QD-Score and coverage metrics take into account redundancies, while their "unique" versions only take into account redundant solutions once.}
\label{tab:stats}
\end{center}
\end{table*}

\subsection{Results}

Experimental cases are compared using the previously defined metrics across 20 runs. We consider all cases listed in Table~\ref{tab:cases}.
Table~\ref{tab:stats} regroups most of these statistics at the last iteration of all cases.

\subsubsection{Training and Shared solution strategies}
We will first compare the performances of the "Base" cases listed in Table~\ref{tab:cases}. 

Figure~\ref{fig:res1} shows representative examples of the four grids found by the \textbf{hardcoded-4}, \textbf{reco-4} and \textbf{qt-reco-4} cases. The last two grids of the \textbf{hardcoded-4} looks symetrical, suggesting that their FD (joint angles between hips and knees) are heavily correlated. The \textbf{reco-4} FD spaces all follow normal distributions, with a large number of inaccessible bins. The \textbf{qt-reco-4} occupy nearly all possible bins.

The statistics in Table~\ref{tab:stats} indicate that the human-designed FD are heavily correlated both in the \textbf{hardcoded-4} and \textbf{hardcoded-4-ns} cases. This translates into container redundancy above $35\%$. The large difference in both coverage and QD-score between \textbf{hardcoded-4} and \textbf{hardcoded-4-ns} suggests than the "Non-sharing" strategy can negatively affect both performance and diversity when the FD can heavily correlated.

%The \textbf{pt-reco-4} is particular because it is the case with the higher redundancy, however, it does not exhibit a high FD correlation, suggesting that most the redundant solutions come from the initialisation phase of the algorithm. 
The online case \textbf{reco-4} is superior to the pre-trained case \textbf{pt-reco-4} in most beneficial statistics: it has a higher Unique QD-Score and Unique Coverage, less redundancy, and less FD correlations. Similarly, cases with Quantile transform post-processing are superior to cases without post-processing.

Both the \textbf{qt-reco-4} and \textbf{qt-reco-4-ns} cases outperform the cases with human-designed FD, possibly because the Quantile transform maximise the number of bins that are occupied. However, this translates into less pressure toward competition, explaining why the best-performing individuals have a lower fitness than the best-performing solutions found by the \textbf{hardcoded-4} and \textbf{hardcoded-4-ns} case. Having a higher redundancy (which would suggest more competition between solutions) does not necessarily translate into higher best fitness scores, as shown by case \textbf{pt-reco-4}. 

The \textbf{qt-reco-4-ns} is the "Base" case with the highest Unique QD-Score and Unique coverage, but the lowest best-fitness. We postulate that this is still related to solution competition because of the higher number of unique bins, and it might be alleviated by increasing the total budget of evaluations.

All the online cases display a far larger variability in term of (Unique) QD-score, coverage and FD correlation than the human-designed FD cases and pre-trained cases, suggesting that solution diversity is heavily affected by the resulting auto-encoder models.

All these results are corroborated in Fig.~\ref{fig:resPlots} which shows a comparison of the "Base" cases across a number of metrics and with respect to iterations. Note that the Unique QD-Score and Unique Coverage metrics show that several cases, especially cases with a "Shared solutions" strategy, may have a spike in value just after re-training. However, this spike may quickly degrade as good solutions propagate into several containers, increasing redundancy. This is one advantage of cases with a "Non-shared solutions" strategy, that do not have any easily perceived spike after re-training.

\subsubsection{Impact of number of containers}
We will now compare cases with differing number of containers, as listed in the "Nr. of Grids" section of Table~\ref{tab:cases}. 

The \textbf{hardcoded-1} case has lower QD-Score and coverage scores compared to \textbf{hardcoded-4}, possibly because its FD pair translates into the sparsest grid, as seen in Fig.~\ref{fig:res1}. However, this results in more competition, which explains its high best-fitness score.

The \textbf{qt-reco-1} case only has one container, and such, does not have any redundancy. This explains why it has a higher Unique QD-score than \textbf{qt-reco-4-ns}, even through its QD-Score is lower than the latter. It is also the case with the higher variability of (Unique) QD-Score suggesting that, on setting with only one container, solution diversity is heavily dependent on the kind of representation learning by the auto-encoder.

In most scores, \textbf{qt-reco-9-ns} outperforms \textbf{qt-reco-6-ns} which outperforms \textbf{qt-reco-4-ns} which outperforms \textbf{qt-reco-25-ns}, suggesting that the increasing the number of container is beneficial, up to a point (here around $9$ containers, and lower than $25$). It is also especially relevant to decrease the variability in (Unique) QD-Score and coverage, with \textbf{qt-reco-25-ns} having the lowest variability over these statistics.

\subsubsection{impact of AE loss}
Here, we compare cases modular auto-encoder models training with a loss function with a diversity component. All cases training in such a manner outperform the \textbf{qt-reco-4-ns} case in most scores, except with the Best-fitness scores which are mostly the same. The best scores are obtained by the \textbf{qt-cmd-4-ns} and \textbf{qt-outputs-4-ns} cases, with similar Unique QD-Scores, Unique coverage scores and redundancy. 

The case \textbf{qt-covmin-4-ns}, designed to increase the covariance between FD, displays a higher FD correlation than \textbf{qt-covmax-4-ns}. Surprisingly, it also shows a lower FD correlation score than \textbf{qt-reco-4-ns}, possibly due to the diversity component of the loss function helping the gradient-descent process.

\section{Discussion and Conclusions}
In this paper, we presented \textbf{MC-AURORA}, a variant of the AURORA algorithm able to automatically find several distinct and complementary behavioural characterisations of the solutions of a problem in an unsupervised manner. While AURORA provided this capability to single-container setups with relatively low dimensional FD spaces, we extended this approach to scale to more complex problems that could be represented in a large number of ways, potentially translating into prohibitively large FD spaces.

Our approach relies on ensemble dimensionality reduction techniques, such as Deep Convolutional Auto-Encoders, that are trained on previously encountered solutions to identify the FD of all containers.
QD algorithms usually assume a uniform distribution over the FD space. However, Auto-Encoders latent representations follows normal distributions, resulting in a non-linear amount of solution competition depending on their position in the FD space. We describe a post-processing method to alleviate this problem.

We investigated the interplay between containers: notably, we described how to avoid or make use of solution redundancy over all containers, and how the number of containers could affect performance and diversity. Our results show that tuning the number of containers can have huge impact on the performance and diversity of the solution. Having larger numbers of containers also reduces Unique QD-Score and Coverage variability.

We showed that it is possible to affect the performance and diversity of solutions found with \textbf{MC-AURORA} by training the auto-encoders with a loss function including a diversity component. We tested three types of diversity loss: either based on the reconstructed outputs, or related to the covariance and correlation between FD. Our results suggest than having both a reconstruction and diversity component in the loss function is mostly beneficial compared to just having a reconstruction. However, this is less beneficial than correctly tuning the number of containers, with \textbf{qt-reco-9-ns} having the best overall results in all cases.

Our approach could be improved further in several ways. First, we could use different auto-encoder topologies to cope with highly dimensional observations (\eg LSTM or Transformers). Here, we only used an homogeneous topology for all containers, but several topologies could be used concurrently, to further increase diversity.

Second, we only accounted for simulation stochasticity through a simple explicit averaging scheme. However, recent studies described more complex methods to handle noise with Quality-Diversity algorithms, possibly through adaptive~\cite{justesen2019map} or implicity averaging~\cite{flageat2020fast}, or by using a Variational Auto-encoder~\cite{gaier2020discovering}.

Third, it would be interesting to investigate fine-grained control of solution redundancy across containers, to reduce or dynamically tune their number. In a future paper, we will describe how another post-processing operation commonly used in Machine Learning known as "whitening"~\cite{kessy2018optimal,hahn2019disentangling} could be used to further decorrelate the FD space and reduce the number of redundancies.

%Perspective: whitening as a postprocessing method to decorrelate FD.
%Data-whitening:~\cite{kessy2018optimal}
%Whitening for VAE:~\cite{hahn2019disentangling}
%Disentangling by factorising: ~\cite{kim2018disentangling}

Finally, it may be possible to automatically and dynamically adapt the evaluation budget of the collection of containers by using a multi-container version of ME-MAP-Elites~\cite{cully2020multi} to dynamically give more evaluation budget to containers depending on their statistics such as coverage, QD-score, redundancy, best-ever, etc. This topic will be addressed in a future paper.

\begin{acks}
This work was supported by Grant-in-Aid for JSPS Fellows JP19F19722.

%%  The authors would also like to thank the anonymous referees for
%%  their valuable comments and helpful suggestions. The work is
%%  supported by the \grantsponsor{GS501100001809}{National Natural
%%    Science Foundation of
%%    China}{http://dx.doi.org/10.13039/501100001809} under Grant
%%  No.:~\grantnum{GS501100001809}{61273304}
%%  and~\grantnum[http://www.nnsf.cn/youngscientists]{GS501100001809}{Young
%%    Scientists' Support Program}.
%
\end{acks}

\FloatBarrier
\bibliographystyle{ACM-Reference-Format}
\bibliography{biblio}

\newpage
\appendix

\section{Pairwise KL-Coverage}

We consider another way of comparing together the FD spaces, by using the KL-Coverage metric from~\cite{pere2018unsupervised}, which was also used in the original AURORA paper~\cite{cully2019autonomous}, to check whether two FD spaces are similar to each other (possibly hinting at their representation capabilities). We extend the original definition to handle multi-containers scenarios:
\begin{equation}
KLC = \sum_c^C {\mathcal{D}_{KL}[E_c||A_c]} = \sum_c^C \sum_{i=1}^{10} E_c(i) log(\frac{E_c(i)}{A_c(i)})
\end{equation}
where $C$ is the set of all containers; and for a container $c$: $E_c$ and $A_c$ are respectively the reference and the compared distributions for container $c$. Lower KLC scores indicate more similar FD spaces.

Table~\ref{tab:klc} contains a pairwise comparison of the distribution of all solutions found by each case, by using the KL coverage metric.

\begin{table*}[h]
\begin{center}
%\rotatebox{90}{
%\resizebox{0.95\textheight}{!}{%
\resizebox{0.95\textwidth}{!}{%
% TODO
\begin{tabular}{lllllllllllll}
\hline
                 & hardcoded-4       & hardcoded-4-ns     & pt-reco-4             & qt-reco-4-ns        & hardcoded-1        & qt-reco-1           & qt-reco-9-ns         & qt-reco-25-ns         & qt-outputs-4-ns     & qt-covmin-4-ns      & qt-covmax-4-ns      & qt-cmd-4-ns         \\
\hline
 hardcoded-4     & $3.532 \pm 2.279$ & $37.529 \pm 5.204$ & $237.401 \pm 138.267$ & $67.478 \pm 58.851$ & $45.213 \pm 9.204$ & $15.145 \pm 26.647$ & $110.498 \pm 60.944$ & $381.826 \pm 103.125$ & $33.672 \pm 24.755$ & $41.156 \pm 53.640$ & $88.725 \pm 90.881$ & $64.773 \pm 68.020$ \\
 hardcoded-4-ns  & $2.161 \pm 0.703$ & $6.556 \pm 4.013$  & $217.568 \pm 30.985$  & $60.626 \pm 55.809$ & $28.818 \pm 6.043$ & $13.950 \pm 24.685$ & $110.802 \pm 51.521$ & $372.828 \pm 83.181$  & $29.310 \pm 24.720$ & $41.447 \pm 49.777$ & $81.032 \pm 70.991$ & $60.412 \pm 64.112$ \\
 pt-reco-4       & $3.758 \pm 0.134$ & $4.028 \pm 1.109$  & $261.816 \pm 96.423$  & $47.333 \pm 52.903$ & $23.149 \pm 6.437$ & $15.965 \pm 28.100$ & $123.597 \pm 65.578$ & $394.563 \pm 107.843$ & $34.616 \pm 22.830$ & $32.684 \pm 48.329$ & $69.403 \pm 79.034$ & $51.481 \pm 50.419$ \\
 reco-4          & $3.637 \pm 0.375$ & $3.313 \pm 0.643$  & $223.318 \pm 88.989$  & $41.457 \pm 41.054$ & $17.760 \pm 5.380$ & $18.787 \pm 40.915$ & $128.605 \pm 74.571$ & $372.045 \pm 100.057$ & $31.613 \pm 21.247$ & $31.097 \pm 42.560$ & $55.783 \pm 52.321$ & $53.082 \pm 52.378$ \\
 qt-reco-4       & $3.271 \pm 0.406$ & $3.099 \pm 0.923$  & $267.139 \pm 89.560$  & $51.507 \pm 49.455$ & $18.478 \pm 5.756$ & $17.875 \pm 34.182$ & $135.863 \pm 71.914$ & $407.154 \pm 86.430$  & $31.978 \pm 21.130$ & $26.666 \pm 37.319$ & $60.443 \pm 58.063$ & $50.702 \pm 38.753$ \\
 qt-reco-4-ns    & $2.084 \pm 0.267$ & $1.156 \pm 0.391$  & $220.504 \pm 49.952$  & $61.870 \pm 66.211$ & $21.708 \pm 5.960$ & $13.462 \pm 24.528$ & $125.668 \pm 50.074$ & $402.748 \pm 63.233$  & $29.157 \pm 19.563$ & $29.572 \pm 35.750$ & $67.997 \pm 61.700$ & $44.575 \pm 27.637$ \\
\hline
 hardcoded-1     & $6.486 \pm 1.832$ & $26.617 \pm 7.624$ & $200.256 \pm 65.087$  & $61.268 \pm 52.100$ & $11.683 \pm 7.774$ & $15.867 \pm 27.894$ & $113.038 \pm 63.244$ & $394.578 \pm 95.579$  & $32.772 \pm 26.580$ & $41.489 \pm 56.487$ & $81.368 \pm 69.214$ & $64.164 \pm 72.541$ \\
 qt-reco-1       & $2.901 \pm 0.609$ & $3.005 \pm 1.020$  & $244.099 \pm 100.073$ & $49.318 \pm 45.511$ & $19.863 \pm 8.580$ & $15.005 \pm 28.277$ & $127.464 \pm 67.042$ & $405.576 \pm 66.359$  & $29.161 \pm 19.047$ & $27.656 \pm 37.383$ & $69.335 \pm 70.539$ & $53.770 \pm 46.366$ \\
 qt-reco-6-ns    & $2.099 \pm 0.327$ & $0.893 \pm 0.150$  & $224.688 \pm 46.879$  & $61.226 \pm 61.438$ & $21.702 \pm 5.647$ & $14.487 \pm 26.782$ & $127.950 \pm 54.268$ & $396.338 \pm 59.380$  & $29.915 \pm 20.218$ & $28.606 \pm 35.372$ & $64.242 \pm 60.267$ & $50.063 \pm 34.534$ \\
 qt-reco-9-ns    & $2.199 \pm 0.218$ & $2.716 \pm 1.116$  & $238.029 \pm 86.652$  & $56.751 \pm 51.980$ & $21.300 \pm 5.790$ & $15.079 \pm 27.784$ & $130.979 \pm 70.728$ & $407.644 \pm 68.638$  & $31.054 \pm 20.840$ & $32.416 \pm 43.633$ & $76.785 \pm 80.134$ & $53.509 \pm 44.164$ \\
 qt-reco-25-ns   & $1.976 \pm 0.173$ & $0.863 \pm 0.123$  & $230.240 \pm 52.536$  & $58.067 \pm 53.512$ & $23.484 \pm 5.822$ & $13.652 \pm 24.422$ & $123.643 \pm 51.230$ & $395.059 \pm 64.495$  & $30.882 \pm 20.510$ & $29.721 \pm 39.078$ & $72.903 \pm 74.861$ & $51.196 \pm 36.077$ \\
\hline
 qt-outputs-4-ns & $2.143 \pm 0.417$ & $1.384 \pm 0.469$  & $232.550 \pm 54.277$  & $60.858 \pm 63.674$ & $22.180 \pm 6.173$ & $13.206 \pm 24.088$ & $127.940 \pm 48.767$ & $413.868 \pm 70.623$  & $29.645 \pm 19.025$ & $29.538 \pm 36.087$ & $66.162 \pm 59.543$ & $48.303 \pm 33.298$ \\
 qt-covmin-4-ns  & $2.199 \pm 0.220$ & $1.054 \pm 0.509$  & $224.056 \pm 48.199$  & $61.159 \pm 63.017$ & $25.768 \pm 5.595$ & $13.498 \pm 24.646$ & $126.980 \pm 49.664$ & $407.084 \pm 70.724$  & $31.167 \pm 19.692$ & $32.825 \pm 36.941$ & $68.111 \pm 63.922$ & $47.500 \pm 30.976$ \\
 qt-covmax-4-ns  & $2.175 \pm 0.214$ & $0.928 \pm 0.254$  & $213.556 \pm 49.202$  & $59.329 \pm 60.668$ & $21.273 \pm 5.913$ & $13.184 \pm 24.116$ & $125.341 \pm 52.401$ & $407.264 \pm 66.606$  & $30.564 \pm 19.857$ & $30.020 \pm 35.265$ & $67.472 \pm 63.610$ & $48.234 \pm 32.006$ \\
 qt-cmd-4-ns     & $2.001 \pm 0.233$ & $0.879 \pm 0.264$  & $227.694 \pm 55.555$  & $63.944 \pm 64.977$ & $24.530 \pm 7.232$ & $13.675 \pm 24.866$ & $130.660 \pm 50.915$ & $415.388 \pm 67.833$  & $29.847 \pm 19.153$ & $30.252 \pm 36.339$ & $68.877 \pm 64.653$ & $47.890 \pm 30.716$ \\
\hline
\end{tabular}
}

%}
\caption{Mean pairwise KL-coverage of all studied cases over 20 runs and using 10 bins per dimension: reference distributions are computed over the column cases and tested distributions over the row cases (so each pairwise score is averaged over $20*20=400$ KL-coverage scores). Lower scores mean distributions that are more similar. The numbers following the symbol $\pm$ represent the standard deviation.}
\label{tab:klc}
\end{center}
\end{table*}

The results in Table~\ref{tab:klc} show that the distributions of solutions of pre-trained and online cases are larger than the ones of the human-designed cases (with low pairwise KL-coverage scores when human-designed cases are used to compute reference distributions). The \textbf{pt-reco-4} has a very dissimilar distribution of solutions compared to the other cases, possibly because of its high redundancy.

\end{document}